\PassOptionsToPackage{dvipsnames,table}{xcolor} 
\documentclass[letterpaper]{article} 
\usepackage{aaai2026}  
\usepackage{times}  
\usepackage{helvet}  
\usepackage{courier}  
\usepackage[hyphens]{url}  
\usepackage{graphicx} 
\usepackage{natbib}  
\usepackage{caption} 
\usepackage{tabularx}
\usepackage{amsmath}
\usepackage{amssymb}
\usepackage{amsthm}
\usepackage{booktabs}
\usepackage{multirow}
\usepackage[table]{xcolor}
\DeclareMathOperator{\bigcdot}{\cdot}
\usepackage{subcaption}
\usepackage{bbding}

\usepackage{algorithm}
\usepackage{algorithmic}

\usepackage{newfloat}
\usepackage{listings}
\DeclareCaptionStyle{ruled}{labelfont=normalfont,labelsep=colon,strut=off} 
\floatstyle{ruled}
\newfloat{listing}{tb}{lst}{}
\floatname{listing}{Listing}
\title{Towards Imperceptible JPEG Image Hiding: Multi-Range Representations-Driven Adversarial Stego Generation}
\author{
    Junxue Yang\textsuperscript{\rm 1},
    Xin Liao\textsuperscript{\rm 1},
    Weixuan Tang\textsuperscript{\rm 2},
    Jianhua Yang\textsuperscript{\rm 3},
    Zheng Qin\textsuperscript{\rm 1}\\
}
\affiliations{
    \textsuperscript{\rm 1}College of Computer Science and Electronic Engineering, Hunan University\\
    \textsuperscript{\rm 2}Institute of Artificial Intelligence, Guangzhou University\\
    \textsuperscript{\rm 3}Guangdong Polytechnic Normal University\\

    \{junxueyang, xinliao, zqin\}@hnu.edu.cn, tweix@gzhu.edu.cn, yangjh86@gpnu.edu.cn
}

\usepackage{bibentry}

\begin{document}

\maketitle

\begin{abstract}
	‌Image hiding fully explores the hidden potential of deep learning-based models, aiming to conceal image-level messages within cover images and reveal them from stego images to achieve covert communication. Existing hiding schemes are easily detected by the naked eyes or steganalyzers due to the cover type confined to the spatial domain, single-range feature extraction and attacks, and insufficient loss constraints. To address these issues, we propose a \underline{m}ulti-range \underline{r}epresentations-driven \underline{a}dversarial stego \underline{g}eneration framework called MRAG for JPEG image hiding. This design stems from the fact that steganalyzers typically combine local-range and global-range information to better capture hidden traces. Specifically, MRAG integrates the local-range characteristic of the convolution and the global-range modeling of the transformer. Meanwhile, a features angle-norm disentanglement loss is designed to launch multi-range representations-driven feature-level adversarial attacks. It computes the adversarial loss between covers and stegos based on the surrogate steganalyzer’s classified features, i.e., the features before the last fully connected layer. Under the dual constraints of features angle and norm, MRAG can delicately encode the concatenation of cover and secret into subtle adversarial perturbations from local and global ranges relevant to steganalysis. Therefore, the resulting stego can achieve visual and steganalysis imperceptibility. Moreover, coarse-grained and fine-grained frequency decomposition operations are devised to transform the input, introducing multi-grained information. Extensive experiments demonstrate that MRAG can achieve state-of-the-art performance.
\end{abstract}


\section{Introduction}

\begin{figure}[t]
	\centering
	\includegraphics[width=1.0\columnwidth]{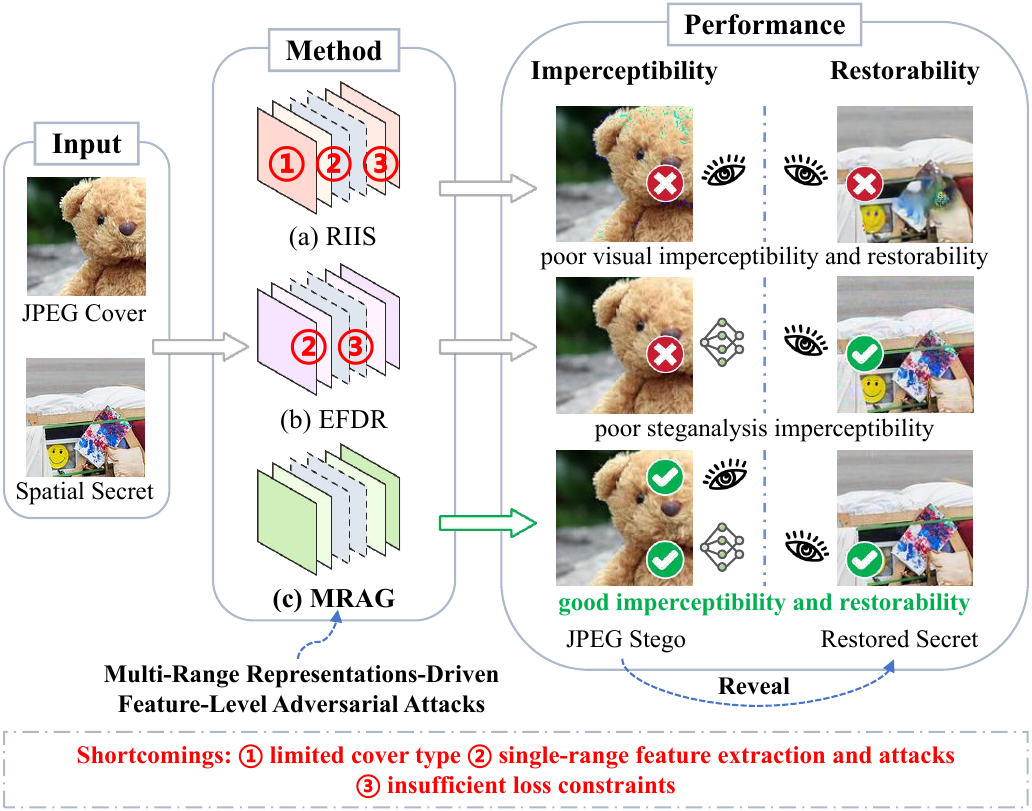} 
	\caption{The illustration shows the performance distinctions between our MRAG and existing image hiding methods when the cover is a color JPEG image. (a) A robust SIH method, RIIS \protect\cite{Xu:22}, produces visible artifacts in the JPEG stego and the restored secret, resulting in poor visual imperceptibility and restorability. (b) Advanced EFDR \protect\cite{Yangb:23} has poor steganalysis imperceptibility. (c) MRAG has satisfying secret restorability, visual imperceptibility, and steganalysis imperceptibility.}
	\label{fig1}
\end{figure}

Image steganography is the art and science of sheltering confidential information in publicly available cover images to obtain stego ones. The resulting stegos should be as unsuspicious as possible, from which secret can be revealed \cite{Provos:03}. It typically has three primary metrics \cite{Pevny:08}: the capacity regarding embedded payloads, the visual imperceptibility by the naked eyes, and the steganalysis imperceptibility against the steganalyzers. The steganalyzer is a binary classifier that distinguishes stego images from cover images. Traditional steganographic methods and earlier deep learning-based methods maintain the bit-level payloads, typically below 0.4 bits per pixel (BPP). The traditional can reach 1 BPP at most, while the deep learning-based can go as high as 4.5 BPP \cite{Lu:21}. In light of this, Baluja \cite{Baluja:17} explores the potential of deep learning-based steganographic models, successfully embedding a spatial RGB image into another of the same size, known as image hiding. Three natural advantages of image hiding are also pointed out: 1) Since the secret is an image with high inherent correlations, the difficulty of network learning can be reduced. 2) The secret itself has a certain fault tolerance. 3) The payload can reach 24 BPP with a 1:1 ratio between the secret and cover. The success of this pioneering work, along with three natural advantages, facilitates the flourishing development of image hiding.

The act of hiding inevitably yields embedded artifacts. Whether artifacts are discovered usually depends on three factors \cite{Baluja:20}: the hidden capacity, the imperceptibility of the algorithm, and the texture of the image itself. The likelihood of detection is negatively correlated with the latter two factors, but positively correlated with the hidden capacity. In this sense, the capacity advantage of image hiding also makes it more vulnerable to detection, turning into a disadvantage. Most hiding methods \cite{Zhang:20,Lu:21,Jing:21,Guan:23,Xu:22} default to such a disadvantage, preferring to improve capacity, visual imperceptibility, and robustness. Only several works \cite{Zheng23,Zhang:21,Hu24} explore the steganalysis imperceptibility utilizing generative adversarial networks (GAN) or adversarial attacks. Due to the inherent instability and convergence difficulties in min-max optimization, GAN-based approaches perform poorer steganalysis imperceptibility. To further improve steganalysis imperceptibility, adversarial attacks-based methods typically employ a pre-trained surrogate steganalyzer and introduce an adversarial loss. Nevertheless, the cover type of the aforementioned methods is exclusively lossless spatial RGB images. We term these methods as spatial image hiding (SIH) for brevity. When the cover is a color JPEG image, the ubiquitous lossy image format, SIH methods perform poorly. There are visible artifacts in the generated JPEG stego and the restored secret. Even the robust SIH method that resists JPEG compression, RIIS \cite{Xu:22}, also encounters the same problem, as shown in Fig.\ref{fig1} (a).

Additionally, the existing SIH methods are restricted to single-domain feature extraction and attacks, lacking steganalysis domain knowledge. The adversarial loss is typically computed as cross-entropy loss based on the surrogate steganalyzer's output. The recent EFDR \cite{Yangb:23} is currently the only work that uses color JPEG images as the covers. Although EFDR can effectively mitigate visible artifacts, its steganalysis imperceptibility remains unsatisfactory due to single-range feature extraction and insufficient loss constraints, as depicted in Fig.\ref{fig1} (b). In summary, existing hiding schemes are easily detected by the naked eyes or steganalyzers due to the cover type confined to the spatial domain, single-range feature extraction and attacks, and insufficient loss constraints. To address these issues, in this paper, we attempt to develop a novel adversarial stego generation framework for color JPEG image hiding, exploring the potential of the generated color JPEG stego in terms of secret restorability, visual imperceptibility, and steganalysis imperceptibility, as illustrated in Fig.\ref{fig1} (c).

For the steganalyzer, to better capture subtle changes, it is essential to consider both local-range and global-range information \cite{Wei:24}. This, in turn, inspires us to construct a \underline{m}ulti-range \underline{r}epresentations-driven \underline{a}dversarial stego \underline{g}eneration framework called MRAG from a steganalysis perspective. MRAG integrates the local-range characteristic of the convolution and the global-range modeling of the transformer. Specifically, MRAG comprises four key components: the invertible local-range branch, the invertible global-range branch, the adaptive fusion module, and the surrogate steganalyzer. A features angle-norm disentanglement loss is incorporated to launch multi-range representations-driven feature-level adversarial attacks. It computes the adversarial loss between covers and generated stegos based on the surrogate steganalyzer’s classified features. Therefore, MRAG is constrained to delicately encode the concatenation of cover and secret into subtle adversarial perturbations from local and global perspectives relevant to steganalysis. We also tailor the coarse-grained frequency decomposition (CFD) and the fine-grained frequency decomposition (FFD) to transform the input. According to the dimension of the obtained transformed images, the transformed images are used as the input for the local and global branches, respectively, naturally introducing multi-grained frequency information. Our contributions can be summarized as follows:

\begin{itemize}
	\item We propose a novel adversarial stego generation framework called MRAG for color JPEG image hiding. MRAG integrates the local-range characteristic of the convolution and the global-range modeling of the transformer from a steganalysis perspective. To our best knowledge, this is the first attempt to explore the potential of the generated color JPEG stego in terms of secret restorability, visual imperceptibility, and steganalysis imperceptibility.
	\item We design a features angle-norm disentanglement loss based on the surrogate steganalyzer’s classified features, which is incorporated to launch multi-range representations-driven feature-level adversarial attacks. Under the dual constraints of features angle and norm, MRAG can delicately encode the concatenation of cover and secret into subtle adversarial perturbations from local and global ranges relevant to steganalysis.
	\item Comprehensive experiments validate the effectiveness and superiority of MRAG on three widely-used datasets, demonstrating that MRAG establishes a new state-of-the-art benchmark in color JPEG image hiding.
\end{itemize}

\section{Related Works}

In this section, we summarize advancements in image hiding and introduce features angle-norm disentanglement.

\begin{figure*}[t]
	\centering
	{\includegraphics[width=1.0\textwidth]{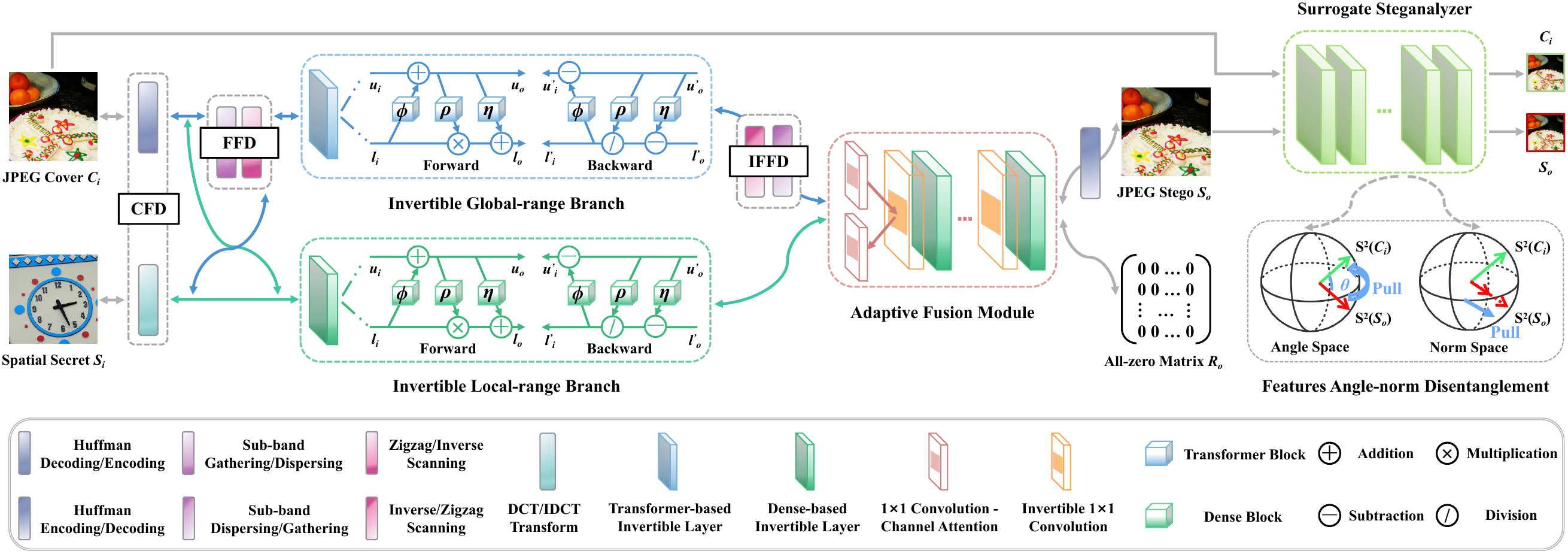}}
	\caption{Overall architecture. To reduce information loss, we utilize invertible neural networks, the mainstream learning paradigm for image hiding, to construct MRAG to the greatest extent. The other image conversion operations and network structures are also reversible and symmetric. In the MRAG's forward-hiding process, the spatial RGB secret image $S_{i}$ is hidden in the color JPEG cover image $C_{i}$ through the invertible local-range branch, the invertible global-range branch, and the adaptive fusion module to obtain the JPEG stego image $S_{o}$ and the redundant information $R_{o}$, an all-zero matrix. In the MRAG's backward-revealing process, $S_{o}$ and $R_{o}$ are fed to reveal the secret image. Features angle-norm disentanglement disentangles the classified features learned by the surrogate steganalyzer. For effectively fooling the surrogate steganalyzer, we pull together the angles between the classified features of cover $\textbf{S}^{2}(C_{i})$ and stego $\textbf{S}^{2}(S_{o})$ while keeping their norm values changeless.}
	\label{fig2}
\end{figure*}

\subsection{Image Hiding} 

Seminal work \cite{Baluja:17} designs an image hiding model containing a secret pre-processing module, a hiding module, and a revealing module. UDH \cite{Zhang:20} is a cover-agnostic image hiding model that discards the pre-processing structure and solely encodes secret images. ISN \cite{Lu:21} treats embedding and extraction as an inverse problem and utilizes the forward and backward processes of invertible neural networks to address it. To improve visual imperceptibility, HiNet \cite{Jing:21} combines invertible neural networks with the wavelet domain and introduces low-frequency wavelet loss. DeepMIH \cite{Guan:23} proposes a novel framework for multiple image hiding. It designs an importance map module to guide the concealing process of the next secret image. CAIS \cite{Zheng23} devises a composition estimation network as the discriminator and promotes steganalysis imperceptibility through generative adversarial training. Both USAP \cite{Zhang:21} and DIH-OAIN \cite{Hu24} utilize adversarial attacks to boost steganalysis imperceptibility. RIIS \cite{Xu:22} is a robust image hiding method that considers various distortions. Nevertheless, the aforementioned works have common shortcomings. The cover type is always lossless spatial RGB images. Their model design lacks steganalysis domain knowledge, relying solely on single-domain feature extraction and attacks. Additionally, the adversarial loss is typically computed as cross-entropy loss based on the surrogate steganalyzer's output. EFDR \cite{Yangb:23} is the first to research image hiding in color JPEG images, focusing on mining fine-grained DCT representations using the transformer operator. Although EFDR has favorable visual imperceptibility and secret restorability, its steganalysis imperceptibility remains poor due to single-range feature extraction and insufficient loss constraints. Noted that, in this paper, we also perform image hiding in color JPEG images, the ubiquitous lossy image format. This is the first attempt to explore the potential of the generated color JPEG stego in terms of secret restorability, visual imperceptibility, and steganalysis imperceptibility.

\subsection{Features Angle-Norm Disentanglement} 

References \cite{Liu:18,chen:20} unveil that the features before the last fully connected layer of the classifier, abbreviated as classified features, can naturally disentangle into angle and norm distances related to inter-class semantic differences and intra-class confidence, respectively. Specifically, they train a plain CNN classifier on the MNIST dataset. Then set the output dimension of classified features to 2 and visualize these features. Three key observations emerge from the visualization. 1) A well-trained classifier can effectively cluster features belonging to the same category. 2) Features angle accounts for semantic/label differences. 3) Features norm reflects intra-class variations. Inspired by these facts, we introduce features angle-norm disentanglement to compute adversarial loss, which pulls together the angles between the classified features of cover and stego while keeping their norm values changeless.

\section{Proposed Framework}

\subsection{Overview}

The pipeline of MRAG consists of four key components: the invertible local-range branch, the invertible global-range branch, the adaptive fusion module, and the surrogate steganalyzer, as shown in Fig. \ref{fig2}. We construct MRAG utilizing invertible neural networks, the mainstream learning paradigm for image hiding, to ensure reversibility as much as possible. The other image conversion operations and network structures are also reversible and symmetric. Therefore, we can implement hiding and revealing utilizing MRAG's forward and backward processes.

The inputs of MRAG's forward-hiding process are the color JPEG cover image $C_{i}$ and the spatial RGB secret image $S_{i}$. Conversion operations contain the coarse-grained frequency decomposition (CFD), the fine-grained frequency decomposition (FFD), and the inverse FFD (IFFD). The inputs are converted into coarse-grained and fine-grained frequency maps through CFD and CFD$\to$FFD, respectively. According to the dimension characteristic of the obtained frequency maps, they are fed into the local and global branches' forward processes, respectively. The dimensions of coarse-grained and fine-grained frequency maps are $(3\times2,H,W)$ and $(8\times8\times3\times2,\frac{H}{8},\frac{W}{8})$ respectively. The superscript denotes the shape with $(channel, height, width)$. The dimension of fine-grained frequency maps is better suited for the patch-wise architecture of the transformer. Then, the global branch's output passes through the IFFD and is concatenated with the local branch's output. Afterward, the concatenation is fed into the adaptive fusion module's forward process to obtain the JPEG stego image $S_{o}$ and the redundant information $R_{o}$, an all-zero matrix. MRAG's backward-revealing process takes the color JPEG stego image $S_{o}$ and $R_{o}$ as input to reveal the secret image. For effectively inducing the surrogate steganalyzer to misclassify the cover $C_{i}$ and stego $S_{o}$, we conduct the features angle-norm disentanglement and pull together the angles between the classified features of cover $\textbf{S}^{2}(C_{i})$ and the classified features of stego $\textbf{S}^{2}(S_{o})$ while keeping their norm values changeless. 

\subsection{MRAG Model}

\paragraph{Backbone structure.} Given that invertible neural networks, the mainstream learning paradigm for image hiding, can preserve as much detail as possible due to their theoretical information lossless characteristic \cite{Dinh:14,Dinh:17,Gilbert:17}, we utilize them to construct our MRAG to the greatest extent. As shown in Fig. \ref{fig2}, key components of MRAG, including the invertible local-range branch, the invertible global-range branch, and the adaptive fusion module, employ invertible neural networks. For the affine coupling layer, the basic unit of invertible neural networks, the input needs to be divided into two parts, $u_{i}$ and $l_{i}$, along the channel axis. Following the reference \cite{Xiao:20}, we use an additive affine transformation for the upper part, $u_{i}$, and employ an enhanced affine transformation for the lower part, $l_{i}$. To be specific, the forward process of the affine coupling layer can be formulated as
\begin{equation}
	\begin{gathered}	
		u_{o}=u_{i}+\phi(l_{i}) \\
		l_{o}=l_{i}\odot{\textbf{exp}(\rho(u_{o}))}+\eta(u_{o})
		\label{eq1}
	\end{gathered}
\end{equation}
In Eq. (\ref{eq1}), $\phi(\bigcdot)$, $\rho(\bigcdot)$, and $\eta(\bigcdot)$ can be arbitrary network structures. $\odot$ and $\textbf{exp}(\bigcdot)$ are the Hadamard product and Exponential function, respectively. Note that $\textbf{exp}(\bigcdot)$ is omitted in Fig. \ref{fig2}. Correspondingly, the inverse process is
\begin{equation}
	\begin{gathered}
		l^{\prime}_{i}=(l^{\prime}_{o}-\eta(u^{\prime}_{o}))\odot\textbf{exp}(-\rho(u^{\prime}_{o})) \\
		u^{\prime}_{i}=u^{\prime}_{o}-\phi(l^{\prime}_{i})
		\label{eq2}
	\end{gathered}
\end{equation}

In the following, we will provide a detailed description of the four key components of MRAG. 

\paragraph{Invertible local-range branch.} As we know, convolution performs excellently in image-related tasks and has nice local neighbor reception characteristics \cite{Krizhevsky:12}. Here, in the invertible local-range branch, $\phi(\bigcdot)$, $\rho(\bigcdot)$, and $\eta(\bigcdot)$ are implemented by a classical 5-layer dense block with a $3\times3$ convolution kernel \cite{Huang:17}. In the dense block, the output feature maps of each layer are passed to subsequent layers, greatly enhancing representation sharing and learning. For the invertible local-range branch, the input is coarse-grained frequency maps through CFD. CFD includes the Huffman decoding of the color JPEG cover image and the DCT transformation of the spatial RGB secret image. 

\paragraph{Invertible global-range branch.} To further enhance representation and global dependency modeling, $\phi(\bigcdot)$, $\rho(\bigcdot)$, and $\eta(\bigcdot)$ in the invertible global-range branch adopt a standard 1-layer Pre-LN transformer \cite{Dosovitskiy:21}. The Pre-LN transformer applies the layer normalization before the multi-head self-attention and the multi-layer perceptron for a more stable gradient distribution. For the invertible global-range branch, the input is the fine-grained frequency maps obtained through CFD$\to$FFD. FFD includes the sub-band gathering and the zigzag scanning. The sub-band gathering rearranges coarse-grained frequency maps, placing the frequency components of the same sub-band on the same channel, making frequency components more compact. The zigzag scanning further rearranges them along the channel axis, grouping similar frequency sub-bands and exploiting the structural information in the JPEG transformation.

\paragraph{Adaptive fusion module.} In addition, to enhance the interaction between local and global representations extracted by the aforementioned two branches, the proposed adaptive fusion module consists of a pair of $1\times1$ convolutions with channel attention and a series of affine coupling layers with an invertible $1\times1$ convolution.  $\phi(\bigcdot)$, $\rho(\bigcdot)$, and $\eta(\bigcdot)$ in the adaptive fusion module also employ the classical 5-layer dense block. The $1\times1$ convolution with channel attention, designed in the reference \cite{Tan:22}, is used to extract the most informative frequency representations adaptively. Adding an invertible $1\times1$ convolution before each affine coupling layer can increase the dependencies between representations, enabling more effective information fusion. For the adaptive fusion module, the input is the concatenation of representations extracted from the invertible local-range branch and representations extracted from the invertible global-range branch after undergoing IFFD. IFFD is used for dimension alignment.

\paragraph{Surrogate steganalyzer.} The surrogate steganalyzer in MRAG can be any well-trained JPEG steganalyzer. Here, we select three open-source steganalyzers: UCNet \cite{Wei:22}, EWNet \cite{Su:21}, and DBS2Net \cite{Hu:24}. UCNet is the sole steganalyzer tailored for color JPEG images. The others are straightforwardly based on existing steganalyzers for gray-scale JPEG images or some well-known computer vision models. EWNet and DBS2Net are gray-scale JPEG steganalyzers. Here, following the setting in the reference \cite{Wei:22}, we adjust the number of input channels in the first convolution layer from 1 to 3, enabling them to detect color images.

\begin{table*}[]
	\centering
	\setlength{\tabcolsep}{0.63mm}
	\begin{tabularx}{\textwidth}{ccccccccccccc}
		\toprule
		\multirow{2.5}{*}{Method} &
		\multicolumn{4}{c}{COCO} &
		\multicolumn{4}{c}{ImageNet} &
		\multicolumn{4}{c}{BOSSBase} \\
		\cmidrule(lr){2-5}\cmidrule(lr){6-9}\cmidrule(lr){10-13}
		\multicolumn{1}{c}{}
		& PSNR$\uparrow$ & SSIM$\uparrow$ & APD$\downarrow$ & \multicolumn{1}{c}{LPIPS$\downarrow$} &  PSNR$\uparrow$ & SSIM$\uparrow$ & APD$\downarrow$ & \multicolumn{1}{c}{LPIPS$\downarrow$} &  PSNR$\uparrow$ & SSIM$\uparrow$ & APD$\downarrow$ & LPIPS$\downarrow$ \\ \midrule
		ISN+ \cite{Lu:21} & 19.85 & .662 & 17.52 & .344 & 19.81 & .643 & 18.98 & .361 & 22.55 & .731 & 14.60 & .345 \\
		RIIS \cite{Xu:22} & 24.98 & .798 & 9.28 & .169 & 24.97 & .781 & 9.80 & .178 & 29.20 & .864 & 5.84 & .143 \\
		EFDR \cite{Yangb:23} & 33.95 & .918 & 4.31 & .028 & 34.05 & .919 & 4.37 & .027 & 36.43 & .921 & 3.22 & .032 \\
		\cellcolor{blue!7} MRAG (UCNet) & \cellcolor{blue!7}40.11 & \cellcolor{blue!7}.983 & \cellcolor{blue!7}1.88 & \cellcolor{blue!7}.004 & \cellcolor{blue!7}39.85 & \cellcolor{blue!7}.980 & \cellcolor{blue!7}2.04 & \cellcolor{blue!7}.004 & \cellcolor{blue!7}41.54 & \cellcolor{blue!7}.975  & \cellcolor{blue!7}1.66 & \cellcolor{blue!7}.005 \\ 
		\cellcolor{blue!7} MRAG (EWNet) & \cellcolor{blue!7}41.67 & \cellcolor{blue!7}.988 & \cellcolor{blue!7}1.62 & \cellcolor{blue!7}.001 & \cellcolor{blue!7}41.20 & \cellcolor{blue!7}.985 & \cellcolor{blue!7}1.80 & \cellcolor{blue!7}.001 & \cellcolor{blue!7}42.32 & \cellcolor{blue!7}.978 & \cellcolor{blue!7}1.55 & \cellcolor{blue!7}.002 \\ 
		\cellcolor{blue!7} MRAG (DBS2Net) & \cellcolor{blue!7}37.14 & \cellcolor{blue!7}.978 & \cellcolor{blue!7}2.78 & \cellcolor{blue!7}.006 & \cellcolor{blue!7}37.00 & \cellcolor{blue!7}.973 & \cellcolor{blue!7}2.90 & \cellcolor{blue!7}.005 & \cellcolor{blue!7}39.49 & \cellcolor{blue!7}.969 & \cellcolor{blue!7}2.15 & \cellcolor{blue!7}.007 \\
		\bottomrule
	\end{tabularx}
	\caption{Secret restorability numerical comparisons. MRAG significantly outperforms existing models.}
	\label{tab:SR}
\end{table*}

\begin{table*}[tbh]
	\centering
	\setlength{\tabcolsep}{0.63mm}
	\begin{tabularx}{\textwidth}{cccccccccccccc}
		\toprule
		\multirow{2.5}{*}{Method} &
		\multicolumn{4}{c}{COCO} &
		\multicolumn{4}{c}{ImageNet} &
		\multicolumn{4}{c}{BOSSBase} \\
		\cmidrule(lr){2-5}\cmidrule(lr){6-9}\cmidrule(lr){10-13}
		\multicolumn{1}{c}{}
		& PSNR$\uparrow$ & SSIM$\uparrow$ & APD$\downarrow$ & LPIPS$\downarrow$ & PSNR$\uparrow$ & SSIM$\uparrow$ & APD$\downarrow$ & LPIPS$\downarrow$ &  PSNR$\uparrow$ & SSIM$\uparrow$ & APD$\downarrow$ & LPIPS$\downarrow$ \\ \midrule
		ISN+ \cite{Lu:21} & 24.09 & .828 & 9.96 & .208 & 24.08 & .823 & 10.66 & .213 & 27.12 & .830 & 7.80 & .239 \\
		RIIS \cite{Xu:22} & 28.25 & .897 & 5.63 & .098 & 28.56 & .895 & 5.86 & .104 & 32.76 & .921 & 3.93 & .102 \\
		EFDR \cite{Yangb:23} & 32.22 & .927 & 4.71 & .025 & 32.04 & .924 & 4.93 & .027 & 36.01 & .944 & 3.12 & .027 \\
		\cellcolor{blue!7} MRAG (UCNet) & \cellcolor{blue!7}31.68 & \cellcolor{blue!7}.920 & \cellcolor{blue!7}5.01 & \cellcolor{blue!7}.028 & \cellcolor{blue!7}31.53 & \cellcolor{blue!7}.916 & \cellcolor{blue!7}5.24 & \cellcolor{blue!7}.031 & \cellcolor{blue!7}35.60 & \cellcolor{blue!7}.940  & \cellcolor{blue!7}3.27 & \cellcolor{blue!7}.031 \\ 
		\cellcolor{blue!7} MRAG (EWNet) & \cellcolor{blue!7}32.78 & \cellcolor{blue!7}.936 & \cellcolor{blue!7}4.39 & \cellcolor{blue!7}.023 & \cellcolor{blue!7}32.62 & \cellcolor{blue!7}.934 & \cellcolor{blue!7}4.58 & \cellcolor{blue!7}.025 & \cellcolor{blue!7}36.37 & \cellcolor{blue!7}.950 & \cellcolor{blue!7}2.98 & \cellcolor{blue!7}.026 \\ 
		\cellcolor{blue!7} MRAG (DBS2Net) & \cellcolor{blue!7}32.35 & \cellcolor{blue!7}.930 & \cellcolor{blue!7}4.65 & \cellcolor{blue!7}.022 & \cellcolor{blue!7}32.16 & \cellcolor{blue!7}.927 & \cellcolor{blue!7}4.86 & \cellcolor{blue!7}.024 & \cellcolor{blue!7}35.92 & \cellcolor{blue!7}.945 & \cellcolor{blue!7}3.16 & \cellcolor{blue!7}.026 \\
		\bottomrule
	\end{tabularx}
	\caption{Visual imperceptibility numerical comparisons. MRAG and EFDR have comparable performance.}
	\label{tab:VI}
\end{table*}

\begin{figure}[]
	\centering
	{\includegraphics[width=0.96\columnwidth]{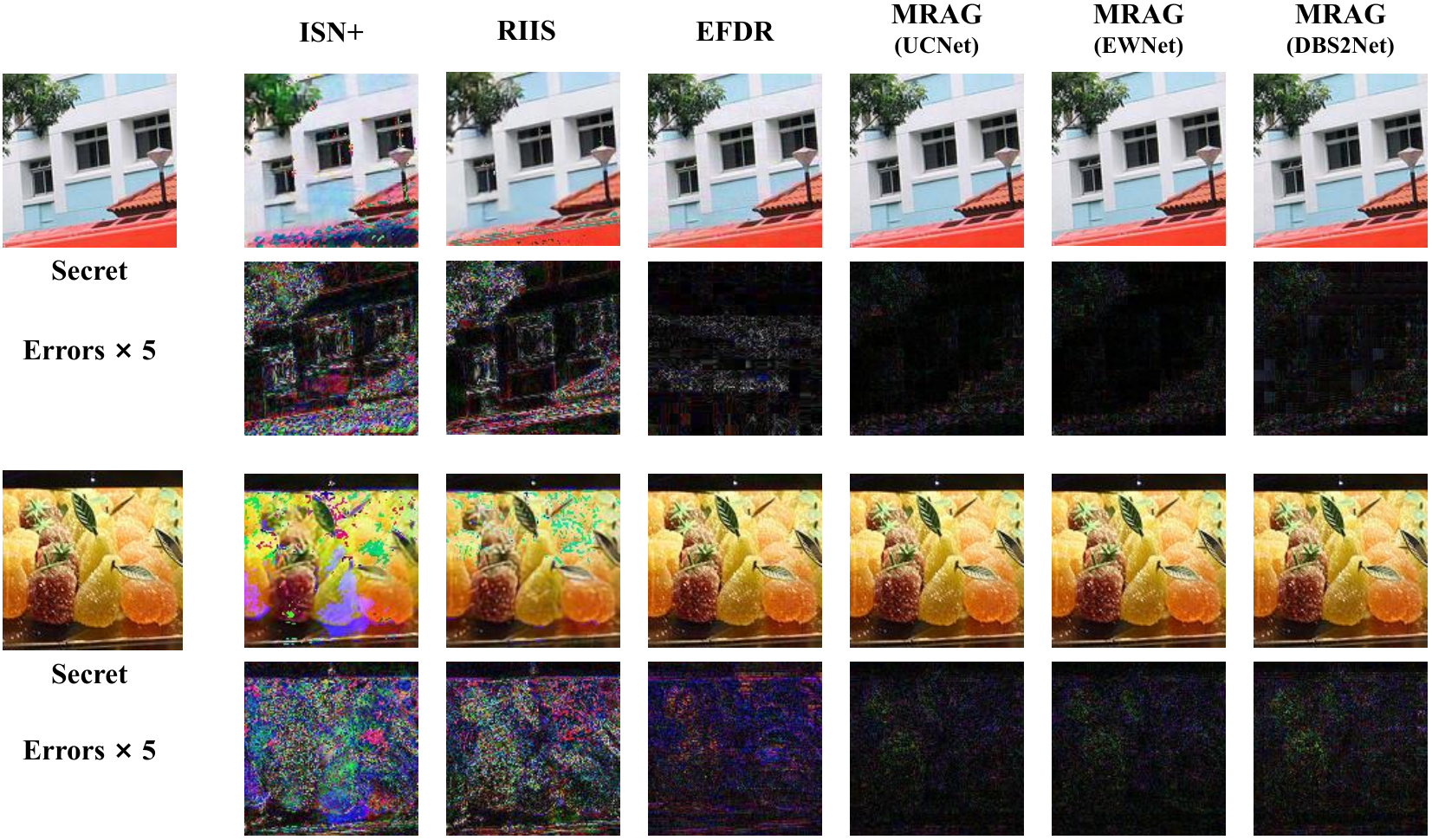}}
	\caption{Secret restorability visualization comparisons. Secrets revealed by MRAG are with minimal visual artifacts.}
	\label{fig3}
\end{figure}

\begin{figure}[t]
	\centering
	{\includegraphics[width=0.96\columnwidth]{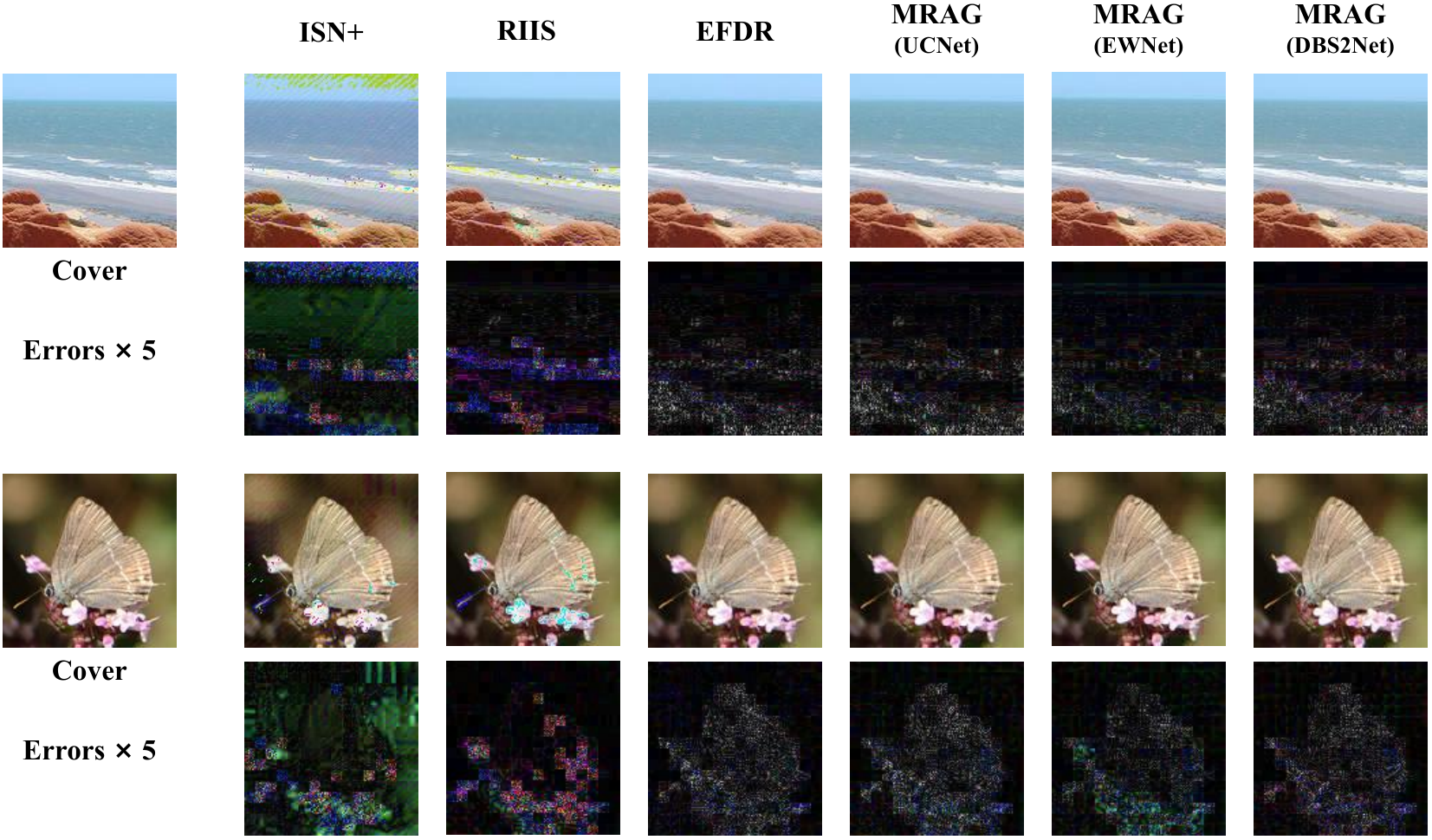}}
	\caption{Visual imperceptibility visualization comparisons. Stegos generated by MRAG are visually friendly.}
	\label{fig4}
\end{figure}

\subsection{Loss Function}
\paragraph{Features angle-norm disentanglement loss.} For fooling the surrogate steganalyzer, we focus on pulling together the angles between the classified features of cover $\textbf{S}^{2}(C_{i})$ and the classified features of stego $\textbf{S}^{2}(S_{o})$. Additionally, to minimize the adversarial perturbations, we keep the norm values of $\textbf{S}^{2}(C_{i})$ and $\textbf{S}^{2}(S_{o})$ changeless. Thus, we can define the features angle-norm disentanglement loss $\mathcal{L}_{an}$ as follows
\begin{equation}
	\begin{gathered}	
		\mathcal{L}_{an}=-\mathcal{L}_{a}+\mathcal{L}_{n} \\
		\mathcal{L}_{a}=CS(\textbf{S}^{2}(C_{i}), \textbf{S}^{2}(S_{o})) \\
		\mathcal{L}_{n}=MSE(\textbf{S}^{2}(C_{i}), \textbf{S}^{2}(S_{o}))
		\label{eq3}
	\end{gathered}
\end{equation}
where $CS$ indicates the cosine similarity. $CS$ measures the similarity between two feature vectors by the cosine value of the angle between them. A larger value indicates a smaller angle, hence a higher similarity. $MSE$ is the mean squared error. $\mathcal{L}_{a}$ and $\mathcal{L}_{n}$ denote the angle loss and norm loss of the classified features, respectively.
\paragraph {Total loss.} The total loss $\mathcal{L}_{total}$ consists of three parts: the hiding loss $\mathcal{L}_{hi}$ to ensure visual imperceptibility, the revealing loss $\mathcal{L}_{re}$ to ensure secret restorability, and the features angle-norm disentanglement loss $\mathcal{L}_{an}$ to guarantee steganalysis imperceptibility. $\mathcal{L}_{total}$ is the sum of these three losses, expressed as follows
\begin{equation}
	\begin{gathered}	
		\mathcal{L}_{total}=\mathcal{L}_{hi}+\mathcal{L}_{re}+\mathcal{L}_{an}
		\label{eq4}
	\end{gathered}
\end{equation}
where ${L}_{hi}=MSE(C_{i}, S_{o})$, and ${L}_{re}=MSE(S_{i}, S^{\prime}_{i})$. $S^{\prime}_{i}$ represents the actual restored secret image.

\begin{table*}[tbh]
	\centering
	\setlength{\tabcolsep}{0.96mm}
	\begin{tabularx}{\textwidth}{ccccccccccccc}
		\toprule
		\multirow{2.5}{*}{Method} &
		\multicolumn{4}{c}{COCO} &
		\multicolumn{4}{c}{ImageNet} &
		\multicolumn{4}{c}{BOSSBase} \\
		\cmidrule(lr){2-5}\cmidrule(lr){6-9}\cmidrule(lr){10-13}
		\multicolumn{1}{c}{}
		& A\_U$\downarrow$ & A\_E$\downarrow$ & A\_D$\downarrow$ & Average$\downarrow$ & A\_U$\downarrow$ & A\_E$\downarrow$ & A\_D$\downarrow$ & Average$\downarrow$ & A\_U$\downarrow$ & A\_E$\downarrow$ & A\_D$\downarrow$ & Average$\downarrow$ \\ \midrule
		EFDR \cite{Yangb:23} & 92.73 & 75.68 & 99.90 & 89.44 & 92.85 & 75.65 & 99.90 & 89.47 & 87.65 & 73.70 & 99.98 & 87.11 \\
		\cellcolor{blue!7} MRAG (UCNet) & \cellcolor{blue!7}50.05 & \cellcolor{blue!7}51.28 & \cellcolor{blue!7}74.53 & \cellcolor{blue!7}58.62 & \cellcolor{blue!7}50.00 & \cellcolor{blue!7}51.68 & \cellcolor{blue!7}78.83 &\cellcolor{blue!7}60.17 & \cellcolor{blue!7}50.00 & \cellcolor{blue!7}50.55 & \cellcolor{blue!7}63.28 & \cellcolor{blue!7}54.61 \\ 
		\cellcolor{blue!7} MRAG (EWNet) & \cellcolor{blue!7}58.53 & \cellcolor{blue!7}60.48 & \cellcolor{blue!7}71.60 & \cellcolor{blue!7}63.54 & \cellcolor{blue!7}60.73 & \cellcolor{blue!7}59.25 & \cellcolor{blue!7}74.65 & \cellcolor{blue!7}64.88 & \cellcolor{blue!7}51.20 & \cellcolor{blue!7}55.00 & \cellcolor{blue!7}59.05 & \cellcolor{blue!7}55.08 \\ 
		\cellcolor{blue!7} MRAG (DBS2Net) & \cellcolor{blue!7}53.35 & \cellcolor{blue!7}51.88 & \cellcolor{blue!7}50.13 & \cellcolor{blue!7}51.79 & \cellcolor{blue!7}53.50 & \cellcolor{blue!7}51.18 & \cellcolor{blue!7}50.30 & \cellcolor{blue!7}51.66 & \cellcolor{blue!7}51.25 & \cellcolor{blue!7}50.43 & \cellcolor{blue!7}50.13 & \cellcolor{blue!7}50.60 \\
		\bottomrule
	\end{tabularx}
	\caption{Steganalysis imperceptibility comparisons. Most of MRAG's classification results approach 50\%, i.e., a random guess. MRAG still maintains higher steganalysis imperceptibility even across the steganalyzer.}
	\label{tab:SI}
\end{table*}

\begin{figure*}[htbp]
	\centering
	\subfloat[UCNet\_EFDR]{\includegraphics[width=0.15\textwidth]{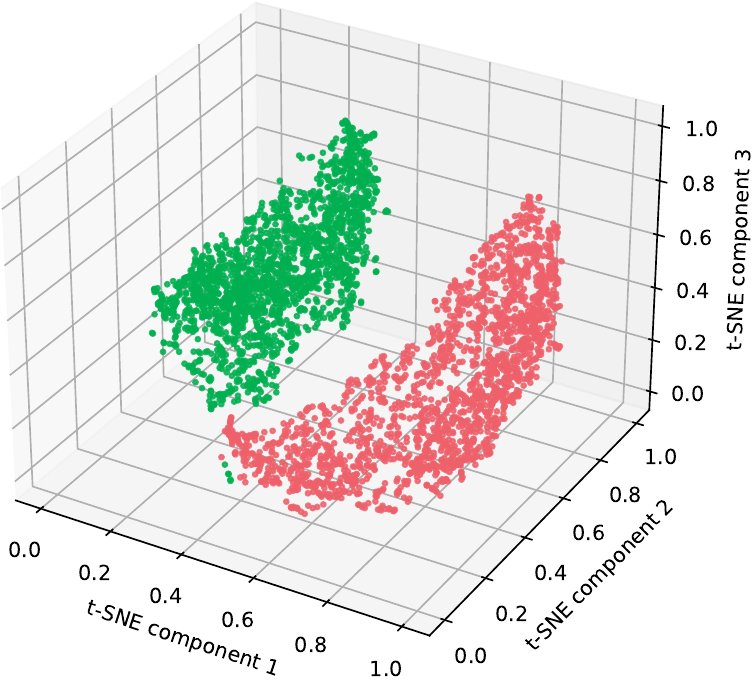}\label{subfig1}}\hfill
	\subfloat[UCNet\_MRAG]{\includegraphics[width=0.15\textwidth]{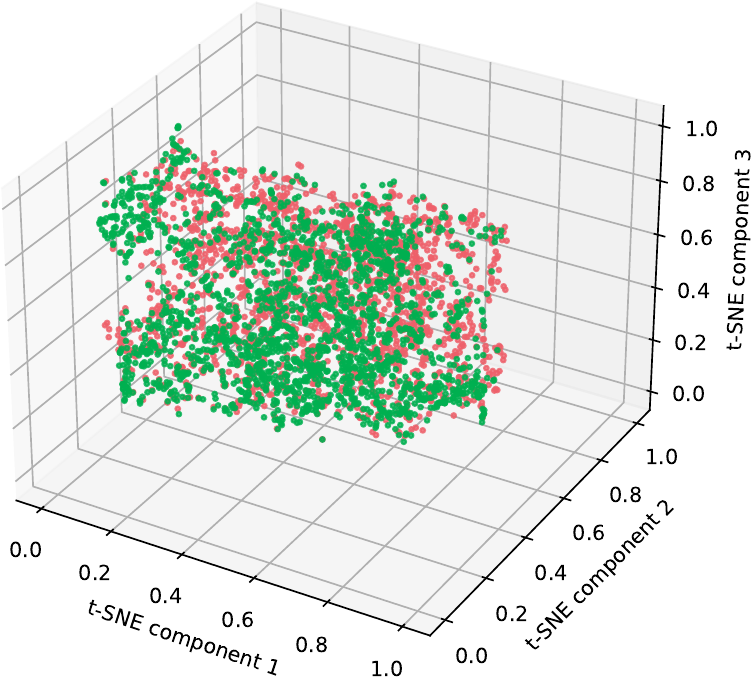}\label{subfig2}}\hfill
	\subfloat[EWNet\_EFDR]{\includegraphics[width=0.15\textwidth]{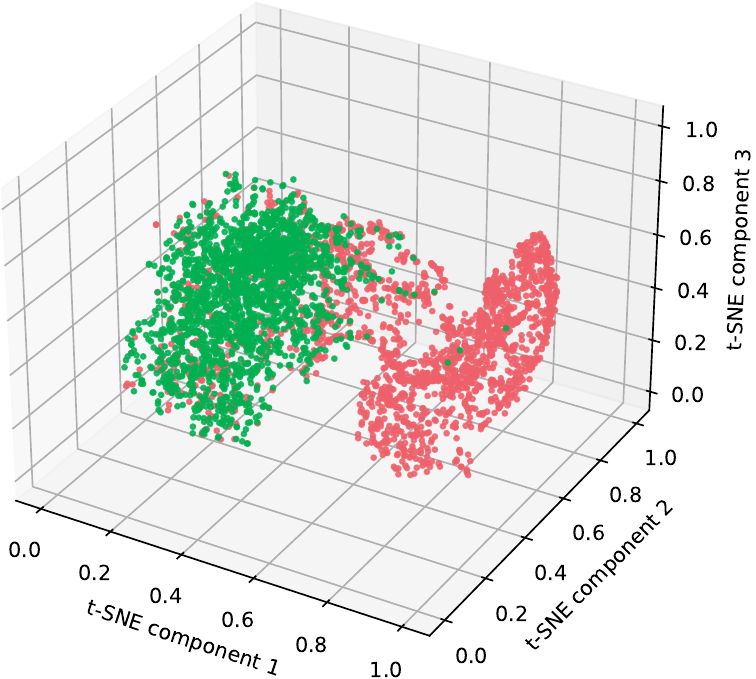}\label{subfig3}}\hfill
	\subfloat[EWNet\_MRAG]{\includegraphics[width=0.15\textwidth]{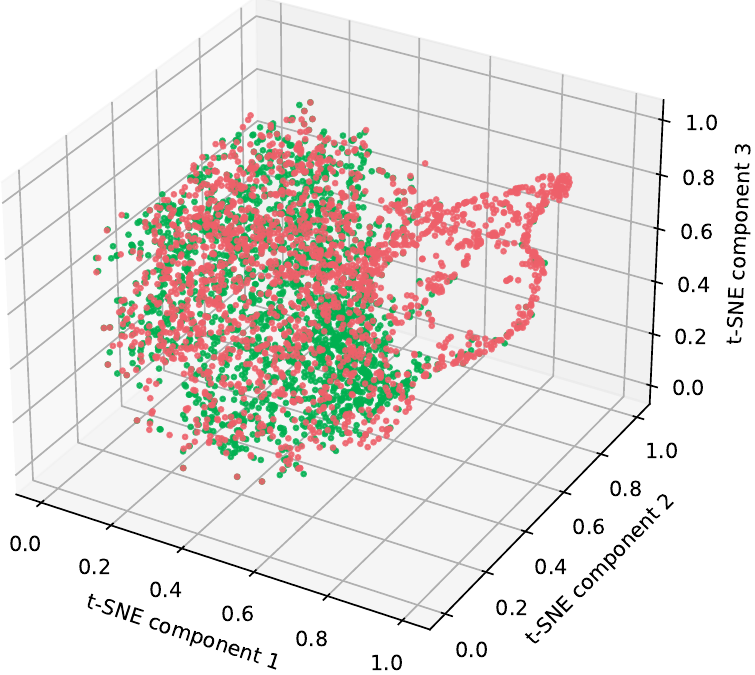}\label{subfig4}}\hfill
	\subfloat[DBS2Net\_EFDR]{\includegraphics[width=0.15\textwidth]{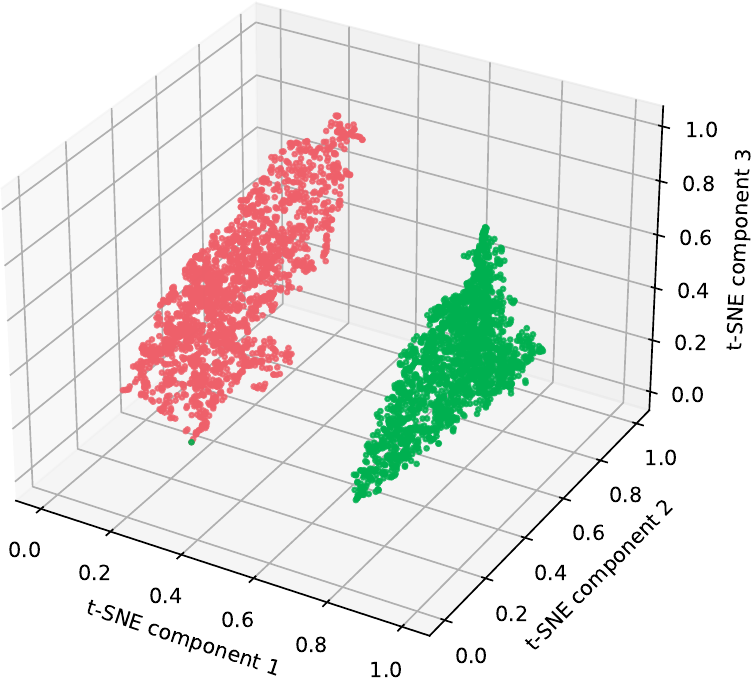}\label{subfig5}}\hfill
	\subfloat[DBS2Net\_MRAG]{\includegraphics[width=0.15\textwidth]{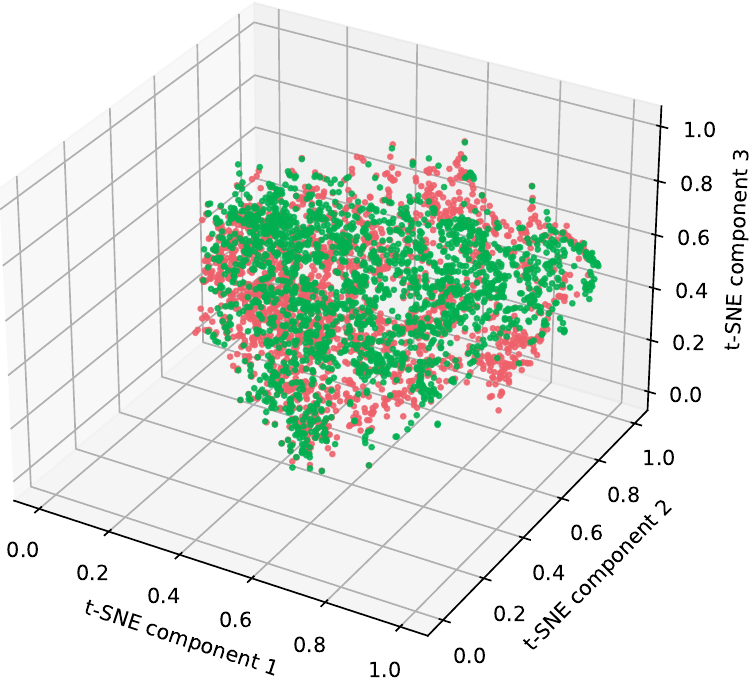}\label{subfig6}}\hfill
	\caption{The t-SNE visualization comparisons of the distribution of the classified features learned by three steganalyzers. Green points denote the distribution of covers' classified features, and red points show the distribution of stegos' classified features. The greater the distinction between two distributions, the easier they are detected by steganalyzers.}
	\label{fig:tnse}
\end{figure*}

\section{Experiments}

\subsection{Experimental Settings}

\paragraph{Datasets.} We employ three image datasets in our implementation, including COCO \cite{Lin:14}, ImageNet \cite{Russakovsky:15}, which are widely used in various deep learning tasks, and a standard steganographic dataset, BOSSBase \cite{Bas:11}. We utilize 5000, 1000, and 2000 cover-secret pairs from COCO for training, validation, and testing, respectively. 2000 cover-secret pairs, each from ImageNet and BOSSBase, are used only for testing. Additional 10000 cover-stego pairs, randomly selected from three datasets, are used to train the steganalyzer. The stego images are generated by HRJS \cite{Yanga:23}, a bit-level JPEG steganographic method based on deep learning. All images are center-cropped to $128\times128$ and $256\times256$, due to the limit of computational resources. The compression quality factor adopted by the cover is 75, the most commonly used in JPEG steganography. The code will be open-sourced on GitHub upon acceptance.

\paragraph{Baselines.} We use several advanced and representative methods as our baselines, including ISN \cite{Lu:21}, a SIH method, RIIS \cite{Xu:22}, a robust SIH method, and EFDR \cite{Yangb:23} that is currently the only customized image hiding method with color JPEG images as covers. Note that during the training phase of ISN, we add an additional JPEG compression distortion layer, referred to as ISN+ for distinction.

\paragraph{Evaluation metrics.} To measure the secret restorability and visual imperceptibility, we adopt peak signal-to-noise ratio (PSNR), structural similarity index measure (SSIM) \cite{Wang:04}, average pixel discrepancy (APD), and learned perceptual image patch similarity (LPIPS) \cite{Zhanga:18}. Higher PSNR/SSIM values and lower APD/LPIPS indicate higher image quality. For steganalysis imperceptibility, we utilize the detection accuracy rate as the metric. The detection accuracy rate is closer to 50\%, i.e., a random guess, indicating superior performance. A\_U, A\_E, and A\_D denote the detection accuracy rate of UCNet, EWNet, and DBS2Net, respectively.

\paragraph{Implementation details.} Before training MRAG, the surrogate steganalyzer is pre-trained for 50 epochs on 5000 mixed cover-stego pairs following default settings. Subsequently, the pre-trained surrogate steganalyzer’s weights are frozen, and we train MRAG for 200 epochs on 5000 cover-secret pairs from COCO. Note that the steganalyzer used for the steganalysis imperceptibility evaluation is trained on another 5000 mixed cover-stego pairs. We adopt the Adam optimizer with $betas=(0.5, 0.999)$, $eps=1e^{-6}$, and $weight\_decay=5e^{-4}$, and set the learning rate to $5e^{-4}$. The mini-batch size is 4, containing two randomly selected cover-secret pairs. All experiments are implemented with PyTorch on the NVIDIA GeForce RTX 4090 GPU.

\begin{table*}
	\centering
	\setlength{\tabcolsep}{3.53mm}
	\begin{tabularx}{\textwidth}{ccccccccc}
		\toprule
		\multirow{2}{*}{The Number of Layer} & \multicolumn{2}{c}{Secret Restorability} & \multicolumn{2}{c}{Visual Imperceptibility} & \multicolumn{4}{c}{Steganalysis Imperceptibility} \\
		\cmidrule(lr){2-3}\cmidrule(lr){4-5}\cmidrule(lr){6-9}
		\multicolumn{1}{c}{} & PSNR$\uparrow$ & SSIM$\uparrow$ & PSNR$\uparrow$ & SSIM$\uparrow$ & A\_U$\downarrow$ & A\_E$\downarrow$ & A\_D$\downarrow$ & Average$\downarrow$ \\
		\midrule
		1 & 41.59 & .987 & 33.21 & .941 & 50.05 & 52.48 & 75.88 & 59.47 \\
		2 & 40.56 & .984 & 32.60 & .934 & 50.08 & 52.85 & 98.90 & 67.28 \\
		\cellcolor{blue!7}3 &\cellcolor{blue!7}40.11  &\cellcolor{blue!7}.983 &\cellcolor{blue!7}31.68 &\cellcolor{blue!7}.920 &\cellcolor{blue!7}50.05 &\cellcolor{blue!7}51.28 &\cellcolor{blue!7}74.53 &\cellcolor{blue!7}58.62 \\
		4 & 39.72 & .983 & 31.80 & .923 & 50.05 & 64.23 & 80.30 & 64.86 \\
		\bottomrule
	\end{tabularx}
	\caption{Ablation for the impact of the number of the affine coupling layer in the adaptive fusion module on performance.}
	\label{tab:ab_BN}
\end{table*}

\begin{table*}
	\centering
	\setlength{\tabcolsep}{3.88mm}
	\begin{tabularx}{\textwidth}{cccccccccc}
		\toprule
		\multirow{2}{*}{Local} & \multirow{2}{*}{Global} & \multicolumn{2}{c}{Secret Restorability} & \multicolumn{2}{c}{Visual Imperceptibility} & \multicolumn{4}{c}{Steganalysis Imperceptibility} \\
		\cmidrule(lr){3-4}\cmidrule(lr){5-6}\cmidrule(lr){7-10}
		\multicolumn{1}{c}{} & \multicolumn{1}{c}{} & PSNR$\uparrow$ & SSIM$\uparrow$ & PSNR$\uparrow$ & SSIM$\uparrow$ & A\_U$\downarrow$ & A\_E$\downarrow$ & A\_D$\downarrow$ & Average$\downarrow$ \\
		\midrule
		\XSolidBrush & \CheckmarkBold & 26.23 & .786 & 31.95 & .924 & 50.03 & 51.15 & 67.75 & 56.31 \\
		\CheckmarkBold & \XSolidBrush  & 36.20 & .965 & 31.27 & .908 & 50.18 & 51.10 & 85.13 & 62.14 \\
		\rowcolor{blue!7} \CheckmarkBold & \CheckmarkBold & 40.11 & .983 & 31.68 & .920 & 50.05 & 51.28 & 74.53 & 58.62\\ 
		\bottomrule
	\end{tabularx}
	\caption{Ablation of the invertible local-range branch and the invertible global-range branch.}
	\label{tab:ab_LG}
\end{table*}

\subsection{Comparison Results}

To verify the performance of MRAG in terms of secret restorability, visual imperceptibility, and steganalysis imperceptibility, we conduct extensive experiments on three datasets and three open-source JPEG steganalyzers.

\paragraph{Secret restorability.} Table \ref{tab:SR} reports the numerical comparisons of MRAG and baselines. It can be observed that MRAG significantly outperforms these baselines in terms of secret restorability. Compared with EFDR, MRAG can supply at least 2.95 dB PSNR advancement. Similar performance improvements can also be observed in SSIM, APD, and LPIPS. Visualization comparisons are shown in Fig. \ref{fig3}. We can observe that the secret images revealed by MRAG are visually pleasing with minimal visual artifacts. In contrast, the SIH methods ISN+ and RIIS yield visible color distortion, blocking, blurring, and banding artifacts in the revealed secret images. It is noteworthy that although our model only uses the COCO for training, it still provides excellent results on the other two datasets, which indicates that our model has a favorable generalization ability.

\paragraph{Visual imperceptibility.} Table \ref{tab:VI} and Fig. \ref{fig4} report the numerical comparisons and visualization comparisons of our MRAG with ISN+, RIIS, and EFDR in terms of visual imperceptibility. We can observe that MRAG and EFDR have comparable performance, while the stegos generated by ISN+ and RIIS still contain many visible artifacts, making it difficult for them to achieve visual imperceptibility.

\paragraph{Steganalysis imperceptibility.} Steganalysis imperceptibility is a more rigorous imperceptibility metric, evaluating the ability to resist steganalyzers. It achieves classification by distinguishing the changes in statistical features caused by hiding. Table \ref{tab:SI} shows the numerical comparisons of our MRAG and EFDR. We can observe that MRAG significantly outperforms EFDR. Compared to EFDR, MRAG can provide at least 15.2\% improvement, with most classification results approaching 50\%. Moreover, we visualize the distribution of the classified features learned by three steganalyzers using t-distributed stochastic neighbor embedding (t-SNE) in Fig. \ref{fig:tnse}. We can find that the distribution of covers' classified features (green points) and stegos' classified features (red points) obtained by EFDR can be well distinguished, while the distribution of the classified features obtained by the proposed MRAG is intermingled and difficult to distinguish. This also confirms the superior steganalysis imperceptibility of MRAG.

Due to the page limit, comparison results with a size of $256\times256$ can be found in the supplementary materials.

\subsection{Ablation Study}

Here, we mainly discuss the impact of the number of the affine coupling layer and multi-range branches. The image size used in ablation is $128\times128$, unless otherwise specified.

\paragraph{Impact of the number of the affine coupling layer.} For the convenience of discussion, we fix the number of the affine coupling layer in the invertible local-range branch and the invertible global-range branch to 1, and focus on investigating the impact of the number of the affine coupling layer in the adaptive fusion module on performance. As reported in Table \ref{tab:ab_BN}, we can observe that the steganalysis imperceptibility performance, the most concerned metric, is optimal when the number of layers is 3. Thus, in our implementation, the number of the affine coupling layer in the adaptive fusion module defaults to 3 unless otherwise specified.

\paragraph{Effectiveness of multi-range branches.} Also, we verify the performance effects of the invertible local-range branch and the invertible global-range branch, as shown in Table \ref{tab:ab_LG}. It can be observed that compared to any single-range branch, the designed multi-range branches can significantly boost the performance of secret restorability and steganalysis imperceptibility while keeping visual imperceptibility.

More experiments include ablation of features angle-norm disentanglement loss and conventional cross-entropy loss, ablation of coarse/fine-grained maps feeding into local/global branches, and comparisons of computational overhead can be found in the supplementary materials.

\section{Conclusion}
This work is the first attempt to explore the potential of the generated color JPEG stego in terms of secret restorability, visual imperceptibility, and steganalysis imperceptibility. It designs a novel adversarial stego generation framework, MRAG, from a steganalysis perspective. Besides, a features angle-norm disentanglement loss, based on the surrogate steganalyzer’s classified features, is incorporated to launch multi-range representations-driven feature-level adversarial attacks. It constrains MRAG to delicately encode the concatenation of cover and secret into subtle adversarial perturbations from local and global ranges relevant to steganalysis. Comprehensive experiments verify the effectiveness and superiority of MRAG. We believe the success of MRAG can further promote the development of color JPEG image hiding and provide some inspiration for further research on steganalysis imperceptibility.

\bibliography{aaai2026}

\end{document}